\documentclass[10pt,twocolumn,letterpaper]{article}

\usepackage[margin=1in]{geometry}

\usepackage{times}

\usepackage{amsmath,amssymb,mathtools}
\usepackage{graphicx}                 
\usepackage{booktabs,multirow,makecell}
\usepackage{siunitx}                  
\usepackage{xcolor}                   
\usepackage{caption,subcaption}       
\graphicspath{{figures/}}             

\usepackage[numbers,sort&compress]{natbib}
\setcitestyle{square,comma,numbers,sort&compress}

\usepackage{amsthm}
\theoremstyle{plain}

\theoremstyle{remark}

\usepackage[
breaklinks=true,
colorlinks=true,
linkcolor=blue,
citecolor=blue,
urlcolor=blue
]{hyperref}

\title{
	FeatureLens: A Highly Generalizable and Interpretable Framework\\
	for Detecting Adversarial Examples Based on Image Features
}

\author{
	Zhigang Yang\textsuperscript{1} \quad
	Yuan Liu\textsuperscript{1} \quad
	Jiawei Zhang\textsuperscript{1} \quad
	Puning Zhang\textsuperscript{1} \quad
	Xinqiang Ma\textsuperscript{2} \\
	\textsuperscript{1}Chongqing University of Posts and Telecommunications \\
	\textsuperscript{2}Chongqing University of Arts and Sciences \\
	{\tt\small yangzhigang@cqupt.edu.cn} \quad
	{\tt\small s230831011@stu.cqupt.edu.cn} \\
	{\tt\small zhangjw@cqupt.edu.cn} \quad
	{\tt\small zhangpn@cqupt.edu.cn} \quad
	{\tt\small xinqma@cqwu.edu.cn}
}

\date{}

\begin{document}
	
	\maketitle
	
	\begin{abstract}
	Although the remarkable performance of deep neural networks (DNNs) in image classification, their vulnerability to adversarial attacks remains a critical challenge. Most existing detection methods rely on complex and poorly interpretable architectures, which compromise interpretability and generalization. To address this, we propose FeatureLens, a lightweight framework that acts as a lens to scrutinize anomalies in image features. Comprising an Image Feature Extractor (IFE) and shallow classifiers (e.g., SVM, MLP, or XGBoost) with model sizes ranging from 1,000 to 30,000 parameters, FeatureLens achieves high detection accuracy—ranging from 97.8\% to 99.75\% in closed-set evaluation and 86.17\% to 99.6\% in generalization evaluation across FGSM, PGD, C\&W, and DAmageNet attacks, using only 51-dimensional features. By combining strong detection performance with excellent generalization, interpretability, and computational efficiency, FeatureLens offers a practical pathway toward transparent and effective adversarial defense.
	
\end{abstract}
	\section{Introduction}
In recent years, deep neural networks (DNNs) have achieved remarkable success across a broad spectrum of applications, including image recognition, autonomous driving, and cybersecurity. These models have emerged as foundational technologies in high-precision perception and intelligent decision-making systems. Representative architectures such as ResNet~\cite{he2016deep}, YOLO (You Only Look Once)~\cite{redmon2016you}, and ViT (Vision Transformer)~\cite{dosovitskiy2020image} have been extensively adopted in various computer vision tasks, consistently attaining state-of-the-art performance. However, despite their notable accuracy, a growing body of research has highlighted critical deficiencies in the robustness of such models.

DNNs have been shown to be highly vulnerable to adversarial perturbations—subtle input modifications that are imperceptible to humans yet sufficient to mislead model predictions. To examine this vulnerability, numerous attack methods have been proposed, including the Fast Gradient Sign Method (FGSM)~\cite{Goodfellow2014}, Projected Gradient Descent (PGD)~\cite{Madry2017}, and the Carlini \& Wagner (C\&W) attack~\cite{Carlini2017}. The key insight gained from this body of research is that these adversarial examples pose significant threats to the reliability of deep learning systems: they preserve perceptually indistinguishable (or semantically intact) input content while deliberately inducing erroneous predictions. This vulnerability is especially alarming in safety-critical applications—such as autonomous driving, medical diagnosis, and biometric authentication—where consistent and robust decision-making under minor perturbations is paramount.	

Adversarial example detection methods fall into two main categories. The first involves training specialized detectors on raw inputs, which offer strong representational power but suffer from high computational costs, instability, and poor generalization under distribution shifts. The second category leverages the target model's internal representations—such as logits, feature activations, or gradients—to distinguish samples \cite{zheng2023detecting}. These white-box approaches achieve strong performance but lack generalizability and interpretability due to their heavy reliance on model-specific features, leading to overfitting in few-shot or cross-model scenarios and linearly scaling costs that hinder large-scale deployment.

Theoretical studies indicate that adversarial detection faces inherent difficulties similar to robust classification~\cite{Tramer2022}, as it requires capturing subtle perturbations to decision boundaries and local structural changes in feature space. This results in opaque decision logic and highly nonlinear internal representations, compromising interpretability and auditability. Practically, existing methods exhibit limited generalization across attack scenarios, where slight distribution shifts or attack modifications cause significant performance degradation. These theoretical and practical constraints highlight the need for an input-driven, interpretable, and model-agnostic detection framework to ensure robust generalization and transparent decision-making across diverse attack scenarios.

\begin{figure*}[htbp]
	\centering
	\includegraphics[width=0.95\linewidth]{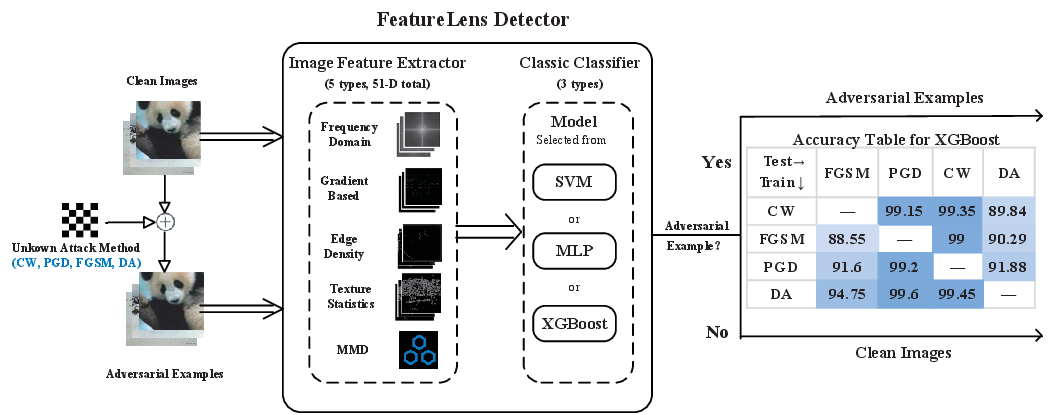}
	\caption{Overview of the proposed FeatureLens framework. Clean and perturbed images—generated by unknown attack methods such as FGSM, PGD, C\&W, and DAmageNet—are processed through a 51-dimensional image feature extractor covering frequency, gradient, edge, texture, and MMD statistics. These image features are classified by a shallow model (SVM, MLP, or XGBoost) to determine whether the input is adversarial. The table on the right presents detection accuracies of XGBoost across cro3ss-attack settings; the complete results are provided in Section~5.3.}
	\label{fig:framework}
\end{figure*}

To address these challenges, we propose FeatureLens—a novel adversarial detection framework emphasizing both generalizability and interpretability. As shown in Figure~\ref{fig:framework}, the proposed framework consists of two components: an Image Feature Extractor (IFE) and a classical classifier. The first component extracts a compact 51-dimensional image vector from the input, while the second employs a shallow classifier to perform detection based on the extracted features. This design ensures that the entire framework remains fully model-agnostic.

Comprehensive evaluations on $\ell_p$-bounded attacks (including FGSM, PGD, and C\&W) and the DAmageNet dataset demonstrate that FeatureLens achieves strong generalization across different attack types. We further assess its performance on a new class of unrestricted attacks—Visual Jailbreak Attacks—where FeatureLens also exhibits robust adaptability beyond norm-constrained scenarios. These results indicate that the proposed architecture, with its compact and semantically transparent features, can effectively differentiate between adversarial and clean samples. This enables interpretable, lightweight, and model-agnostic detection without requiring access to the target model’s internal parameters.

Our contribution in this paper are as follows:

\begin{itemize}
	\item \textbf{Model-agnostic adversarial detection framework.}  
	To address the limitation of existing state-of-the-art detection methods that rely on the internal parameters of target models, we design a black-box detection architecture that does not depend on model states or forward processes, thereby achieving truly model-agnostic adversarial detection. The proposed framework exhibits high deployment flexibility and provides a new implementation pathway for generalized adversarial defense.
	
	\item \textbf{Feature-driven detection mechanism with interpretability and strong generalization.}  
	We propose a feature-driven detection mechanism that combines image features with shallow classifiers (SVM, MLP, and XGBoost). Through a lens-like analysis, it reveals the intrinsic differences between clean and adversarial samples. In across attack evaluations across FGSM, PGD, CW, and DAmageNet attacks, the method maintains high interpretability while demonstrating excellent generalization performance.
	
	\item \textbf{Lightweight deployment and practical system integration.}  
	The proposed detection framework significantly reduces model complexity while maintaining high accuracy. The number of parameters is reduced from the million-level scale of conventional deep detection models (approximately $10^6$) to the ten-thousand-level scale (approximately $10^4$), achieving a reduction of about two orders of magnitude. This lightweight design enables efficient operation across embedded devices, edge computing nodes, and server environments, providing strong real-time performance and resource adaptability.
\end{itemize}
	\section{Related Work}

Adversarial detection research has evolved along two main directions: input-based and model-based methods. The former detects adversarial examples by analyzing perturbation patterns directly from the input image, emphasizing interpretability and generality, while the latter leverages internal representations—such as activations or gradients—for higher detection accuracy. However, balancing interpretability, discriminative power, and generalization remains an open challenge.

\paragraph{Input-Based Detection Paradigm.}

Input-based detection methods identify anomalies by analyzing perturbation patterns in the input image without relying on the target model’s internal representations. These approaches emphasize generality and efficiency by exploiting the intrinsic consistency of input features to reveal perturbation-induced shifts. Representative works include ADDITION~\cite{wang2023addition}, which detects anomalies via adaptive denoising; a pseudo-random encoding–based model-agnostic detector~\cite{zhu2023toward}; ENAD~\cite{craighero2023unity}, integrating Mahalanobis distance, LID, and one-class SVM for stability; and a dual-classifier design leveraging spatial transformation sensitivity~\cite{Tian2021}. Overall, these lightweight, model-agnostic approaches demonstrate the potential of input-feature–driven detection, though the absence of a unified theoretical foundation still limits their robustness under complex distributions.

From the perspective of interpretability, input-based detection methods typically employ shallow classifiers such as SVM, $k$-NN, or MLP, combined with explicit feature design to ensure transparent decision processes. Li et al.~\cite{li2017adversarial} revealed perturbation-induced distributional shifts through convolutional filter analysis. Tian et al.~\cite{tian2018detecting} measured perturbation sensitivity by examining prediction instability under input transformations. Gorbett et al.~\cite{Gorbett2022} demonstrated the linear separability of intermediate representations in shallow feature spaces. These studies suggest that shallow architectures, owing to their simplicity and well-defined decision boundaries, provide high interpretability and traceability. Nevertheless, the enhancement of interpretability often entails a reduction in discriminative power because shallow architectures are limited in capturing high-dimensional nonlinear relationships under complex perturbation conditions, resulting in lower detection accuracy compared with deep counterparts.

To improve robustness and generalization across diverse attack scenarios, several studies have refined the modeling of input-based detection methods by incorporating richer statistical priors or multi-domain features. Abusnaina et al.~\cite{Abusnaina2021} employed graph neural networks to characterize latent-space neighborhood relations for multi-attack detection. Gao et al.~\cite{Gao2023a} adopted a dual-domain autoencoder to evaluate anomaly severity through reconstruction errors. Zhao et al.~\cite{Zhao2024} proposed a robustness score based on minimal perturbation magnitude to unify evaluation across different attack types. These advancements mitigate the reliance on specific attack assumptions and improve across attack transferability. However, inconsistencies in feature normalization and data distribution hinder performance stability, and the balance between generalization robustness and interpretive coherence remains unresolved. In summary, input-based detection combines interpretability with efficiency but still faces limitations in theoretical consistency and cross-distribution stability, motivating exploration of model-based approaches for greater discriminative power and robustness.

\paragraph{Model-Based Detection Paradigm}

In contrast, model-based detection methods analyze perturbation propagation within the internal representations of the attacked model by accessing feature layers, gradient distributions, or output responses. Mosca et al.~\cite{mosca2022suspicious} examined variations in the logits layer to capture perturbation-induced output shifts. Zheng et al.~\cite{zheng2023detecting} quantified gradient sharpness in loss space to evaluate local parameter sensitivity. Zhang et al.~\cite{zhang2024dofa} proposed the DoFA framework, which extracts attribution maps from multiple convolutional layers to localize perturbations. Wang et al.~\cite{wang2025detecting} improved robustness by modeling gradient-noise distributions using pseudo-adversarial data. Jamil et al.~\cite{Jamil2023} investigated binary ReLU activation patterns and inter-layer Hamming similarities to trace abnormal propagation paths in deep architectures. These methods generally achieve high accuracy under white-box conditions but depend heavily on parameter access and gradient computation, resulting in substantial computational costs and limited transferability across models or attack types.

Regarding interpretability, model-based detection methods often employ post-hoc analyses such as saliency visualization, gradient attribution, or attention mapping to elucidate internal response mechanisms. Ye et al.~\cite{ye2022detection} detected anomalies by analyzing inconsistencies between input saliency and attention responses. Wang et al.~\cite{wang2022adversarial} compared multi-layer saliency patterns to trace perturbation propagation within deep feature spaces. Although these methods help in understanding the decision logic of deep detectors, their explanatory stability is undermined by strong dependencies on network structure, gradient information, and noise levels, which hinders the establishment of a consistent and verifiable interpretive framework. Consequently, a pronounced trade-off persists between accuracy and interpretability reliability.

In terms of generalization, model-based detection remains vulnerable to unseen attacks. Carlini and Wagner~\cite{Carlini2017a} circumvented multiple detectors through tailored loss design, revealing dependence on heuristic assumptions. Tramer~\cite{Tramer2022} highlighted that diverse perturbation training can enhance generalization but at a significant computational cost. Karunanayake et al.~\cite{Karunanayake2025} formalized robustness constraints under distributional shifts. While model-based methods exhibit strong representational capacity and high detection accuracy, their stability under unknown or distribution-shifted conditions remains limited, and a unified theoretical understanding of generalization is still lacking. Overall, the tension between precision and transferability constitutes a fundamental bottleneck in this paradigm.

Taken together, input-based and model-based detection paradigms represent two complementary perspectives. The former emphasizes external feature consistency and interpretability, whereas the latter focuses on internal representational sensitivity and detection accuracy. The former demonstrates superior transferability under complex conditions but limited modeling capacity, while the latter achieves higher precision in white-box environments yet lacks generalizability. This study aims to bridge the theoretical divide between these two paradigms by constructing an image feature space that captures the impact of perturbations on feature distributions, thereby achieving efficient, interpretable, and across attack generalizable adversarial detection without reliance on model parameters.
	\section{Methodology}

This section introduces a lightweight and model-agnostic detection framework that identifies adversarial examples using image features extracted directly from input images. Classical classifiers such as SVM, MLP, and XGBoost operate on these features to perform discrimination in a unified input space. We describe the overall formulation, feature construction, and training protocol.

\subsection{Problem Formulation}

The goal of adversarial detection is to identify perturbed inputs without accessing internal model states. Let $\mathcal{X} \subset \mathbb{R}^{H \times W \times 3}$ denote the input space and $\mathcal{Y} = \{0,1\}$ the label space, where $y=1$ indicates an adversarial sample and $y=0$ a normal one.

We define a feature extractor $\phi: \mathcal{X} \rightarrow \mathbb{R}^d$ that maps an image $\mathbf{x}$ to a $d$-dimensional image vector $\mathbf{z} = \phi(\mathbf{x})$, composed of frequency statistics, gradient structure, edge density, texture responses, and a distributional shift score via maximum mean discrepancy (MMD).

Detection is framed as binary classification in the feature space. Given classifier $f: \mathbb{R}^d \rightarrow \mathcal{Y}$, the objective is to minimize expected risk:

\begin{equation}
	\mathcal{L}(f) = \mathbb{E}_{(\mathbf{x}, y) \sim \mathcal{D}} \left[ \ell(f(\phi(\mathbf{x})), y) \right]
\end{equation}

\noindent where $\ell(\cdot, \cdot)$ is the cross-entropy loss and $\mathcal{D}$ the data distribution.

Unlike prior approaches relying on logits or internal activations, our method constructs an interpretable, model-agnostic representation using 51 handcrafted features across four domains: spectral, gradient, edge/texture, and distributional shift. These descriptors are designed to reflect structural anomalies caused by adversarial perturbations and form the foundation for detection.

\subsection{Image Feature Representation}

\paragraph{Frequency-Domain Features}

Adversarial perturbations, though typically imperceptible in the spatial domain, tend to amplify high-frequency components, resulting in measurable distortions in the Fourier spectrum. We perform a 2D Fourier transform and partition the magnitude spectrum into low-frequency, mid-frequency, and high-frequency bands to compute their relative energy ratios. To further characterize high-frequency anomalies, we extract both energy concentration and average magnitude within the high-frequency region. Additionally, spectral entropy, skewness, kurtosis, and contrast are derived to describe the distribution shape and spectral complexity. Collectively, these frequency-domain features provide interpretable and robust indicators of perturbation-induced artifacts, with detailed formulations included in the appendix A.1.

\paragraph{Gradient-Based Structural Features}

Gradient-based features capture local intensity transitions and structural patterns—such as edges, contours, and textures—which are often disrupted by adversarial perturbations. In contrast to pixel-level descriptors, gradient representations exhibit greater sensitivity to fine-grained spatial inconsistencies. We compute per-pixel gradient magnitudes and orientations using Sobel filters, and extract both global statistics—including the mean, standard deviation, and entropy—as well as a 36-bin directional histogram to characterize edge strength and orientation coherence. Complete definitions are provided in Appendix~A.2.

\paragraph{Edge and Texture Features}

Beyond spectral and gradient cues, adversarial perturbations can cause localized geometric distortions and textural inconsistencies. To capture these effects, we introduce two compact descriptors: edge density, which summarizes boundary continuity, and texture response mean, which reflects average directional energy from Gabor filters across multiple frequencies and orientations. These features provide complementary structural priors that enhance sensitivity to localized, interpretable perturbations. Full definitions are provided in Appendix~A.3.

\paragraph{Distributional Shift Feature}

The aforementioned features, while effective at capturing local structure, typically do not encode a global view of a sample’s position within the overall data distribution. To integrate this perspective, we introduce a distributional shift feature based on MMD, measuring a sample’s deviation from a normal reference set in feature space. This MMD score provides a model-agnostic indication of statistical abnormality and enhances local descriptors by exposing distribution-level inconsistencies—such as those found in edge cases or generative outliers. Appended as the 51st dimension of the feature vector, it improves sensitivity to out-of-distribution shifts. Additional details are provided in Appendix A.4.

\paragraph{Feature Fusion and Output Representation}

We integrate all extracted descriptors into a unified 51-dimensional feature vector that captures diverse image features for model-agnostic detection. This semantically organized representation comprises: frequency-domain features (9D) ,  gradient-based features (39D), edge and texture features (2D) , and an MMD (1D). All features are Z-score standardized, combining local descriptors with global alignment into a compact detection signature. Complete details are provided in Appendix A.5.
	\section{Theoretical Analysis and Interpretability Verification}

This section provides a unified examination of the proposed structured image features, demonstrating both their theoretical separability and interpretability. We first analyze the geometric displacement induced by adversarial perturbations to prove linear separability in the feature space, and then assess attribution consistency to validate that the detection mechanism remains transparent, stable, and auditable across models.

\subsection{Linear Separability in Image Feature Space}

Most existing adversarial detection methods rely on internal model states—such as logits, gradients, or hidden activations—thus tightly coupling the detector to the classifier. This limits their applicability in black-box or heterogeneous environments. In contrast, our method leverages only image features extracted directly from the input, enabling detection with shallow, model-agnostic classifiers.

Let $x$ denote a clean input and $x' = x + \eta$ its adversarial counterpart, with $\|\eta\|_2 \leq \varepsilon$. The feature mapping $\phi(x) \in \mathbb{R}^{d}$ encodes interpretable descriptors such as high-frequency energy $f_{\text{HF}}$ and gradient entropy $H_G$. We first establish the following:

\textbf{Theorem 1.} For any $\|\eta\|_2 \leq \varepsilon$, adversarial perturbations lead to measurable displacements in the image feature space:
\begin{equation}
	\|\phi(x') - \phi(x)\|_2^2 \geq \delta_1^2 + \delta_2^2
\end{equation}
where $\delta_1 = |f_{\text{HF}}(x') - f_{\text{HF}}(x)|$ and $\delta_2 = |H_G(x') - H_G(x)|$.

This quantifiable shift supports linear separability.

\textbf{Theorem 2.} If $\|\phi(x') - \phi(x)\|_2 \geq \Delta$, and $\Delta$ exceeds intra-class variation, then there exists a linear classifier $f(x) = w^\top \phi(x) + b$ such that:
\begin{equation}
	\text{sign}(f(x)) \cdot \text{sign}(f(x')) < 0
\end{equation}
A constructive proof (see Appendix~B.1) shows that setting $w := \phi(x') - \phi(x)$ and $b := - w^\top \phi(x) - \frac{1}{2} \|w\|_2^2$ achieves the desired separation.

These results demonstrate that adversarial perturbations cause measurable and interpretable displacements in the image feature space, confirming the geometric separability of adversarial and clean samples. To further understand which specific image features dominate this separation and how they contribute to the detection decision, we next analyze feature attribution and interpretability.

\subsection{Feature Attribution and Interpretability}

Beyond detection accuracy, image features must also support transparent and traceable decision-making—crucial for security auditing, adversarial forensics, and system accountability. We analyze interpretability using XGBoost as a representative shallow model, focusing on three widely adopted feature importance metrics: \textit{gain}, \textit{cover}, and \textit{weight}, which respectively quantify how much a feature improves splits, how many samples it affects, and how frequently it is used.

To complement these global metrics, we further apply SHAP analysis to provide local, sample-level attributions. As detailed in Appendix~B.3, frequency-domain statistics (e.g., \textit{MidFreqRatio}, \textit{FreqEntropy}) and gradient-based image features (e.g., \textit{GradEntropy}, \textit{GradHist\_20}) show the highest SHAP values across samples, aligning closely with the global gain ranking. This consistency demonstrates that the most influential image features are stable across classifiers (XGBoost, SVM, MLP) and form the primary basis for adversarial discrimination.

\begin{figure}[htbp]
	\centering
	\includegraphics[width=0.95\linewidth]{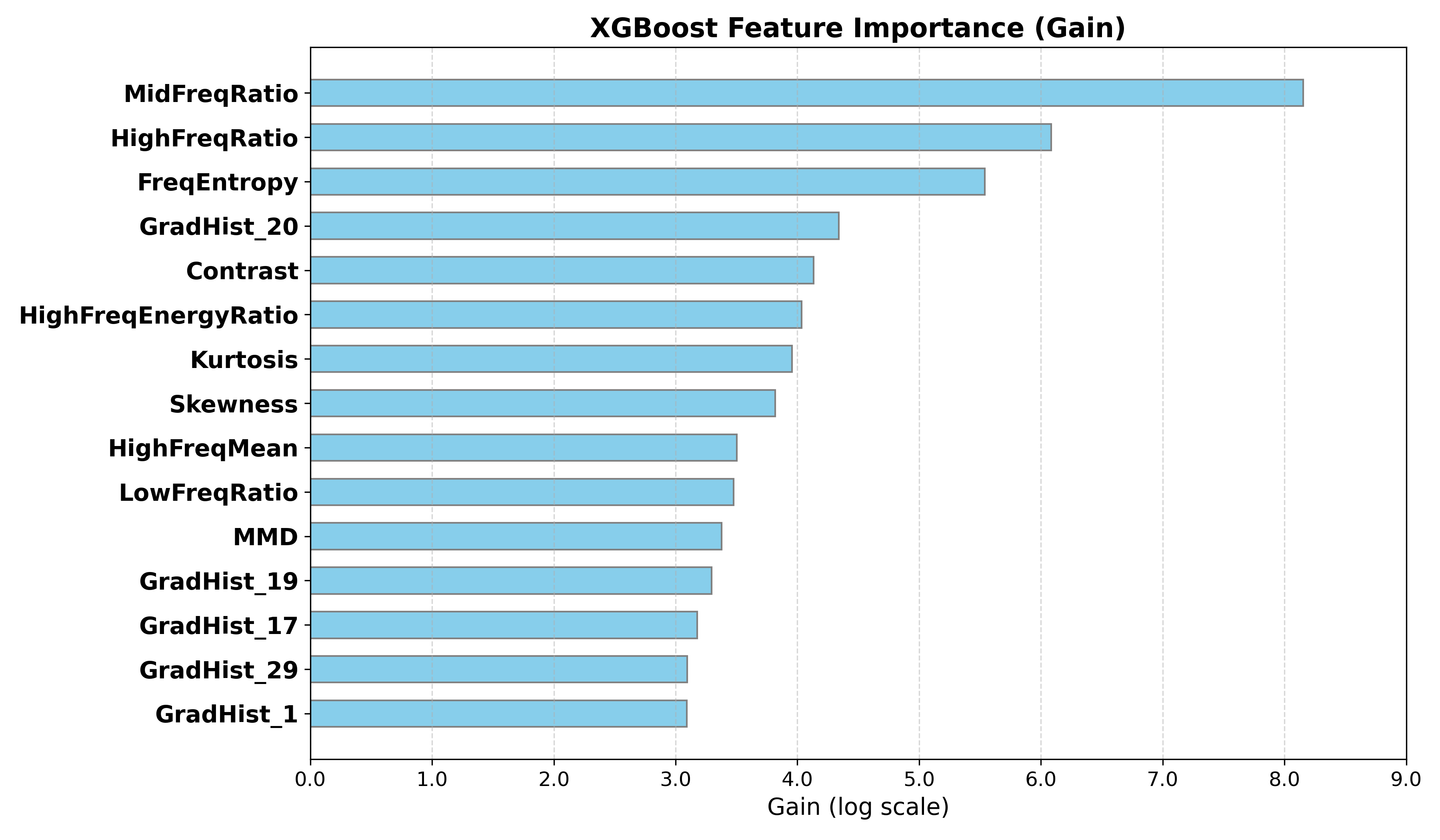}
	\caption{
		XGBoost feature importance results based on the \textit{gain} metric. This metric quantifies the average reduction in the loss function contributed by splits involving each feature, reflecting its overall influence in decision-making. The results show that frequency-domain features (e.g., MidFreqRatio, FreqEntropy) exhibit strong discriminative power across multi-dimensional representations, while gradient histogram features (e.g., GradHist\_20) capture structural and textural information essential for distinguishing perturbation patterns. Their prominence suggests that the model primarily relies on frequency and gradient cues to identify subtle adversarial artifacts, underscoring the core role of structural information in adversarial detection.
	}
	\label{fig:xgb_importance_gain}
\end{figure}

As shown in Figure~\ref{fig:xgb_importance_gain}, the gain-based attribution results highlight the dominant influence of frequency-domain statistics (e.g., MidFreqRatio, FreqEntropy) and orientation-sensitive gradient features (e.g., GradHist\_20) on model decisions. To validate the stability of this trend, supplementary analyses using the \textit{cover} and \textit{weight} metrics are provided in AppendixB.3, offering a more comprehensive perspective.

To validate attribution consistency, we present SHAP and gradient-based analyses in AppendixB.3, demonstrating that frequency and gradient features are consistently prioritized across XGBoost, SVM, and MLP. Building on this insight, we conducted feature dimensionality reduction by removing low-contribution dimensions, resulting in a compact 37-dimensional representation. To assess whether this streamlined representation maintains sufficient detection capacity, we evaluated it on the Hybrid dataset. As shown in AppendixB.3, the reduced feature set leads to a performance drop of approximately 5\%, yet still achieves strong detection results across all classifiers, confirming the robustness and efficiency of the selected features.
	\section{Experiments}

The following presents a comprehensive evaluation of the proposed detection framework across three key scenarios: closed-set detection, across attack generalization, and adaptation to emerging threats such as Visual Jailbreak Attacks. Results demonstrate that image features, combined with shallow classifiers, enable robust, model-agnostic adversarial detection.

\subsection{Experimental Setting}

This section details the experimental setup used to evaluate the effectiveness and generalizability of the proposed image feature-based detection framework.

\paragraph{Pipeline Overview}
The detection pipeline consists of four stages: (1) adversarial sample generation, (2) image feature extraction, (3) shallow classifier training, and (4) standardized evaluation. The framework is fully model-agnostic, requiring no access to deep network internals, thus ensuring portability across architectures.

\paragraph{Attack Types and Generalization Protocol}
We evaluate the detector against three canonical white-box attacks—FGSM, PGD, and C\&W—each generated via gradient-based optimization under different norm constraints. To assess out-of-distribution robustness, we also include DAmageNet, a black-box benchmark composed of naturally adversarial examples from ImageNet. For generalization evaluation, we construct a unified training regime (Hybrid\_24000) combining data from all four attack types.

\paragraph{Dataset Splits}
Each white-box attack dataset contains 6000 samples (3000 adversarial, 3000 normal), split into 60\% training, 20\% validation, and 20\% testing. DAmageNet is excluded from training and reserved solely for testing to ensure statistical independence. All subsets are class-balanced and stratified. Complete dataset statistics are listed in Appendix~C.1.

\paragraph{Feature Extraction and Classifiers}
Each input image is transformed into a 51-dimensional vector, composed of interpretable frequency, gradient, edge, and texture features. All features are standardized via Z-score standardization. We evaluate three representative shallow classifiers: a Gaussian-kernel SVM, a two-layer MLP (64 and 32 hidden units, ReLU activations), and an XGBoost classifier (max depth 6, 100 trees, learning rate 0.1). All models are trained, validated, and tested independently per configuration.

\paragraph{Evaluation Metrics}
Detection performance is measured using accuracy, F1-score, and AUC. Emphasis is placed on generalization and robustness, supported by confusion matrix and ROC curve analyses.

\begin{table}[t]
	\centering
	\scriptsize
	\setlength{\tabcolsep}{3pt}
	\caption{
		Closed-set detection performance of the proposed method and two baselines under matched training and testing conditions across four representative attack types.
		The evaluation uses 6000 training samples per attack.
		For our method, results from three shallow classifiers (SVM, MLP, and XGBoost) trained on image features are reported.
		AaD~\cite{Zhao2024} leverages perturbation cost sensitivity, while HSGL~\cite{Jamil2023} constructs ReLU activation patterns and clusters them via Hamming similarity.
		Each value represents mean Accuracy, F1, and AUC (all in \%).
		Bold values indicate the best performance per attack.
	}
	\label{tab:closedset-full}
	\resizebox{\linewidth}{!}{
		\begin{tabular}{llccc}
			\toprule
			\textbf{Attack} & \textbf{Method} & \textbf{Accuracy} & \textbf{F1} & \textbf{AUC} \\
			\midrule
			\multirow{5}{*}{FGSM}
			& FeatureLens (SVM) & 97.80 & 97.85 & 98.48 \\
			& FeatureLens (MLP) & 99.60 & 99.60 & 99.99 \\
			& FeatureLens (XGB) & \textbf{99.75} & \textbf{99.75} & \textbf{100.00} \\
			& AaD               & 91.75 & 89.29 & 91.02 \\
			& HSGL              & 98.73 & 98.52 & 99.22 \\
			\midrule
			\multirow{5}{*}{PGD}
			& FeatureLens (SVM) & 98.10 & 98.14 & 98.66 \\
			& FeatureLens (MLP) & 99.40 & 99.40 & 99.81 \\
			& FeatureLens (XGB) & \textbf{99.85} & \textbf{99.85} & \textbf{99.93} \\
			& AaD               & 90.10 & 87.73 & 88.86 \\
			& HSGL              & 98.92 & 98.64 & 99.35 \\
			\midrule
			\multirow{5}{*}{C\&W}
			& FeatureLens (SVM) & 98.40 & 98.42 & 99.16 \\
			& FeatureLens (MLP) & 99.50 & 99.50 & 99.93 \\
			& FeatureLens (XGB) & \textbf{99.70} & \textbf{99.70} & \textbf{99.98} \\
			& AaD               & 89.98 & 88.97 & 88.39 \\
			& HSGL              & 98.42 & 98.08 & 98.99 \\
			\midrule
			\multirow{5}{*}{DAmageNet}
			& FeatureLens (SVM) & 98.16 & 98.21 & 98.66 \\
			& FeatureLens (MLP) & 99.62 & 99.63 & 99.86 \\
			& FeatureLens (XGB) & \textbf{99.50} & \textbf{99.50} & \textbf{99.92} \\
			& AaD               & 91.13 & 89.50 & 90.26 \\
			& HSGL              & 98.12 & 97.94 & 98.62 \\
			\bottomrule
		\end{tabular}
	}
\end{table}

\begin{table*}[t]
	\centering
	\scriptsize
	\setlength{\tabcolsep}{3pt}
	\caption{
		Comprehensive generalization evaluations (Accuracy / AUC, in \%) for five detection methods: FeatureLens (SVM), FeatureLens (MLP), FeatureLens (XGB), AaD, and HSGL. Each row indicates the adversarial attack used during training, while each column (excluding the diagonal) denotes a different unseen attack used for testing. The “Hybrid” column aggregates performance when evaluated on the full test set composed of all four attack types. The rightmost “Mean ± Std” column summarizes the average and variability (standard deviation) of generalization performance across all heterogeneous train-test combinations. Diagonal cells (i.e., same train and test attack) are excluded to avoid overestimating intra-distribution performance.
	}
	\label{tab:cross_generalization_all}
	\resizebox{\linewidth}{!}{
		\begin{tabular}{llccccc|c}
			\toprule
			\textbf{Method} & \textbf{Train~$\downarrow$~Test~$\rightarrow$} & \textbf{FGSM} & \textbf{PGD} & \textbf{C\&W} & \textbf{DA} & \textbf{Hybrid} & \textbf{Mean ± Std} \\
			\midrule
			\multirow{4}{*}{\makecell{FeatureLens \\ (SVM)}}
			& FGSM      & — / —       & 98.05 / 98.56 & 97.70 / 99.16 & \textbf{97.08} / \textbf{98.26} & 97.50 / 98.60 & 97.58 ± 0.35 / 98.65 ± 0.33 \\
			& PGD       & 97.85 / 97.96 & — / —       & 98.00 / 98.68 & 97.00 / 98.39 & 96.31 / 98.40 & 97.29 ± 0.68 / 98.36 ± 0.26 \\
			& C\&W      & 98.25 / 98.37 & 98.35 / 98.71 & — / —       & 88.79 / 97.97 & 94.41 / 98.50 & 94.95 ± 3.90 / 98.39 ± 0.27 \\
			& DAmageNet & 97.40 / 98.45 & 98.15 / 98.61 & \textbf{98.00} / \textbf{98.94} & — / —       & 98.10 / 98.70 & 97.91 ± 0.30 / 98.68 ± 0.18 \\
			\cmidrule(lr){1-8}
			\multirow{4}{*}{\makecell{FeatureLens \\ (MLP)}} 
			& FGSM      & — / —       & 98.75 / 99.64 & 91.95 / 97.81 & 89.89 / 99.73 & 94.20 / 99.40 & 93.15 ± 3.29 / 99.15 ± 0.93 \\
			& PGD       & 98.95 / 99.85 & — / —       & 95.75 / 99.36 & 90.43 / 99.63 & 97.60 / 99.10 & 95.18 ± 3.58 / 99.24 ± 0.99 \\
			& C\&W      & 99.00 / 99.79 & 99.30 / 99.89 & — / —       & 86.17 / 99.19 & 93.91 / 99.64 & 94.60 ± 5.31 / 99.63 ± 0.27 \\
			& DAmageNet & 99.25 / 99.86 & 96.05 / 99.68 & 83.95 / 97.08 & — / —       & 95.81 / 99.35 & 93.77 ± 6.46 / 98.99 ± 1.01 \\
			\cmidrule(lr){1-8}
			\multirow{4}{*}{\makecell{FeatureLens \\ (XGB)}} 
			& FGSM      & — / —       & 99.00 / 99.90 & 88.55 / 97.90 & 90.29 / 99.80 & 93.90 / 99.90 & 92.43 ± 4.55 / 99.41 ± 1.04 \\
			& PGD       & 99.20 / 99.90 & — / —       & 91.60 / 99.80 & 91.88 / 99.70 & 96.20 / 99.80 & 94.72 ± 3.57 / 99.55 ± 0.85 \\
			& C\&W      & 99.15 / 99.85 & 99.35 / 99.90 & — / —       & 89.84 / 99.25 & 95.60 / 99.80 & 95.49 ± 4.32 / 99.70 ± 0.33 \\
			& DAmageNet & \textbf{99.60} / \textbf{99.95} & \textbf{99.45} / \textbf{99.90} & 94.75 / 97.25 & — / —       & \textbf{98.50} / \textbf{99.90} & \textbf{98.08 ± 2.01} / \textbf{99.88 ± 0.12} \\
			\midrule
			\multirow{4}{*}{AaD} 
			& FGSM      & — / — & 83.59 / 82.03 & 82.59 / 81.30 & 85.63 / 84.92 & 84.28 / 80.01 & 84.02 ± 1.11 / 82.06 ± 1.80 \\
			& PGD       & 83.59 / 82.03 & — / — & 83.74 / 82.20 & 84.86 / 84.05 & 78.12 / 76.24 & 82.58 ± 2.62 / 81.13 ± 2.93 \\
			& C\&W      & 83.65 / 82.61 & 83.10 / 81.40 & — / — & 85.84 / 85.17 & 84.99 / 84.00 & 84.40 ± 1.08 / 83.30 ± 1.42 \\
			& DAmageNet & 86.93 / 85.72 & 85.08 / 83.84 & 82.06 / 81.40 & — / — & 83.10 / 82.48 & 84.29 ± 1.87 / 83.36 ± 1.61 \\
			\midrule
			\multirow{4}{*}{HSGL} 
			& FGSM      & — / — & 84.06 / 83.25 & 82.14 / 80.43 & 87.04 / 86.42 & 85.27 / 84.62 & 84.60 ± 1.96 / 84.18 ± 1.29 \\
			& PGD       & 84.06 / 83.25 & — / — & 82.38 / 81.11 & 85.20 / 84.06 & 84.92 / 84.39 & 84.39 ± 1.30 / 83.75 ± 1.36 \\
			& C\&W      & 84.13 / 83.17 & 83.21 / 82.19 & — / — & 86.30 / 85.38 & 85.18 / 84.68 & 84.70 ± 1.15 / 83.86 ± 1.25 \\
			& DAmageNet & 87.04 / 86.42 & 85.20 / 84.06 & 82.06 / 82.02 & — / — & 86.59 / 86.15 & 85.72 ± 1.91 / 85.76 ± 1.05 \\
			\bottomrule
		\end{tabular}
	}
\end{table*}

\subsection{Closed-Set Evaluation}

We begin by evaluating the performance of image features under a closed-set setting, where the attack type used during training matches that used for testing. Four representative attack types are considered—FGSM, PGD, C\&W, and DAmageNet—with 6000 samples per attack (3000 adversarial and 3000 normal), evenly split across training, validation, and testing.

Table~\ref{tab:closedset-full} summarizes the results under matched train-test settings for four representative attacks. All three shallow classifiers (SVM, MLP, and XGBoost ) trained on image features achieve high accuracy, with XGBoost performing best overall. C\&W and PGD are effectively detected by all models, while DAmageNet presents more challenges due to its natural perturbations. Nevertheless, our method consistently achieves strong results across metrics, demonstrating robustness and adaptability. Detailed confusion matrices are provided in Appendix~C.2.

For comparison, we evaluate two representative baselines. Attack as Detection (AaD)~\cite{Zhao2024} estimates vulnerability based on the perturbation cost needed to mislead a model, while Hamming Similarity with Graph Laplacian (HSGL)~\cite{Jamil2023} clusters activation patterns via spectral analysis of ReLU-based Hamming graphs. As shown, both baselines perform reasonably on gradient-based attacks but degrade significantly on DAmageNet, underscoring the stronger generalization of model-agnostic image features.

\subsection{Generalization Evaluation}

In practical deployment scenarios, detectors are likely to encounter adversarial perturbations that differ significantly from those observed during training. This distributional mismatch necessitates robust generalization beyond known attacks. To systematically evaluate generalization across attacks, we construct 16 transfer configurations in which the adversarial types used during training and testing are deliberately disjoint, spanning four representative attacks: FGSM, PGD, C\&W, and DAmageNet.

Table~\ref{tab:cross_generalization_all} summarizes the cross-generalization performance of five representative detection methods—FeatureLens (SVM), FeatureLens (MLP), FeatureLens (XGB), AaD, and HSGL—evaluated over all train-test combinations among four adversarial attacks (FGSM, PGD, C\&W, and DAmageNet), as well as a comprehensive Hybrid test set. Diagonal entries where training and testing attacks are the same are excluded.

Our FeatureLens framework with XGBoost achieves the best overall performance (Acc: 95.22\%, AUC: 99.42\%) across all transfer settings. In comparison, AaD and HSGL peak at only 85.72\% and 85.76\% respectively, falling short even of FeatureLens variants using SVM or MLP. This indicates that FeatureLens captures the essential distinctions between adversarial and clean examples, enabling substantial gains in generalization.

Notably, under unknown or mixed attack conditions, FeatureLens (XGB) remains robust, with low variance and consistently strong results across train-test shifts. These findings validate the framework’s practical utility and superior adaptability under real-world threat diversity.

Overall, these results underscore the effectiveness of our model-agnostic image feature representation, which enables consistent and interpretable detection across diverse adversarial domains. The original heatmaps and complete tables illustrating generalization across attacks are presented in Appendix~C.3 for reference.

\subsection{New Attack Detection}

To assess our framework's adaptability to emerging threat modalities, we evaluate it against the Visual Jailbreak Attack~\cite{Qi2024}, which embeds fine-grained perturbations into clean images and pairs them with malicious prompts to induce unauthorized responses in Vision-Language Models. This black-box attack bypasses model internals and poses a unique cross-modal safety challenge, the system framework diagram of the proposed method is included in Appendix~C.3 for reference.

\begin{table}[!htbp]
	\centering
	\caption{Detection accuracy and AUC (\%) of the proposed method on Visual Jailbreak adversarial images.}
	\label{tab:jailbreak-detection}
	\setlength{\tabcolsep}{6pt} 
	\begin{tabular}{p{3.8cm}<{\raggedright}c}
		\toprule
		Attack Type & Acc./AUC (\%) \\
		\midrule
		Adv. ($\epsilon_4{=}16/255$) & 86.82 / 0.930 \\
		Adv. ($\epsilon_4{=}32/255$) & 92.41 / 0.958 \\
		Adv. ($\epsilon_4{=}64/255$) & 94.83 / 0.970 \\
		Adv. (unconstrained) & 98.20 / 0.987 \\
		\bottomrule
	\end{tabular}
\end{table}

Following the original evaluation protocol, we construct a Visual Jailbreak dataset comprising four perturbation levels ($\epsilon_1=16/255$, $\epsilon_2=32/255$, $\epsilon_3=64/255$, and $\epsilon_4$ unconstrained). For each level, 500 clean ImageNet images are paired with 500 corresponding adversarial counterparts, forming a balanced 1:1 ratio. The entire dataset is randomly divided into training and testing sets with an 8:2 split. All samples are processed through the proposed image feature extractor and evaluated using a pre-trained shallow detector.

As shown in Table~\ref{tab:jailbreak-detection}, detection accuracy increases steadily from 86.82\% at $\epsilon_4=16/255$ to 98.20\% under the unconstrained setting. The accompanying AUC values (0.93–0.99) exhibit the same trend, supporting that stronger perturbations yield more distinguishable structural deviations. These results confirm the framework’s robustness in detecting alignment-targeted visual jailbreaks without relying on model-specific cues.
	\section{Conclusion}

This paper presents a lightweight, model-agnostic framework for adversarial example detection based on 51-dimensional image features extracted from frequency, gradient, edge, texture, and MMD, without relying on model internals. Experimental results show that shallow classifiers trained on these features achieve stable and high detection accuracy under closed-set, across attack, and new attack scenarios, demonstrating the effectiveness and adaptability of the proposed feature space against diverse adversarial threats.

Theoretical analysis further confirms that adversarial perturbations induce measurable and structured displacements in the image feature space, leading to geometric separability between clean and adversarial samples. Feature attribution results reveal that frequency- and gradient-based descriptors play dominant roles in detection decisions and remain consistent across classifiers, validating the interpretability, stability, and model-agnostic reliability of the proposed framework.

Future work will extend this feature-based paradigm to other input modalities and multimodal fusion settings, and explore dynamic feature selection and compression strategies to further reduce complexity without compromising robustness. Overall, the findings suggest that interpretable statistical priors provide a promising foundation for building reliable and generalizable adversarial detection systems.
	
	{\small
		\bibliographystyle{ieeenat_fullname}
		\bibliography{main}
	}

\end{document}